# Using UAVs for vehicle tracking and collision risk assessment at intersections

**Shuya Zong**
Graduate Research Assistant, Center for Connected and Automated Transportation (CCAT), and Lyles School of Civil Engineering, Purdue University, West Lafayette, IN, 47907.
Email: szong@purdue.edu

**Sikai Chen***
Visiting Assistant Professor, Center for Connected and Automated Transportation (CCAT), and Lyles School of Civil Engineering, Purdue University, West Lafayette, IN, 47907.
Email: chen1670@purdue.edu; and
Visiting Research Fellow, Robotics Institute, School of Computer Science, Carnegie Mellon University, Pittsburgh, PA, 15213.
Email: sikaichen@cmu.edu
ORCID #: 0000-0002-5931-5619
(Corresponding author)

**Majed Alinizzi**
Assistant Professor, Department of Civil Engineering, College of Engineering, Qassim University, 51452 Buraydah, Qassim, Saudi Arabia
Email: alinizzi@qec.edu.sa

**Yujie Li**
Graduate Research Assistant, Center for Connected and Automated Transportation (CCAT), and Lyles School of Civil Engineering, Purdue University, West Lafayette, IN, 47907.
Email: li2804@purdue.edu
ORCID #: 0000-0002-0656-4603

**Samuel Labi**
Professor, Center for Connected and Automated Transportation (CCAT), and Lyles School of Civil Engineering, Purdue University, West Lafayette, IN, 47907.
Email: labi@purdue.edu
ORCID #: 0000-0001-9830-2071

**ABSTRACT**

Assessing collision risk is a critical challenge to effective traffic safety management. The deployment of unmanned aerial vehicles (UAVs) to address this issue has shown much promise, given their wide visual field and movement flexibility. This research demonstrates the application of UAVs and V2X connectivity to track the movement of road users and assess potential collisions at intersections. The study uses videos captured by UAVs. The proposed method combines deep-learning based tracking algorithms and time-to-collision tasks. The results not only provide beneficial information for vehicle's recognition of potential crashes and motion planning but also provided a valuable tool for urban road agencies and safety management engineers.

## INTRODUCTION

It has been prognosticated that unmanned aerial vehicles (UAVs) will play a vital role in various application or context areas of transportation systems management. This is motivated by the success of UAVs in other domains including photography, photogrammetry, agriculture, terrain mapping, monitoring, disaster relief and rescue operations, and recreational purposes (*1*). Due to these applications, the emerging global market for drone-enabled services has been valued by the 2016 Middle East and North Africa Business Report at over $127B (*2*). Also, it is predicted that the industry will lead to the creation of more than 100,000 new jobs (*3*). According to recent literature (*4*), 7 million small UAVs are already deployed in air space for commercial use in various domains including real states, insurance, and agriculture.

Transportation engineers has investigated various ways in which UAV technology can be applied to enhance transportation operations, and drone-based solutions are being developed and tested to increase the efficiency of transportation in general and freight transportation in particular (*5*). Recognizing the immense potential of UAV technology, the US Congress, in 2012, passed legislation that requires the Federal Aviation Authority (FAA) to integrate small drones into airspace by 2015 (*4*). That legislation propelled UAV research further, and increased the number of research efforts in this area. Most of this research work focused on traffic flow analysis (*6,7*), vehicle detection (*8*) and highway infrastructure management (*9*) while relatively limited attention has been paid to drone-based risk assessment of traffic safety (*4*). Current researches in UAV-based risk assessment are limited. Kim and Chervonenkis studied the detection of emergency and abnormal traffic situations with UAV artificial vision system (*10*). Algorithms developed in their study, however, are limited to recognize only a few abnormal situations. A study from Sharma et al. (*11*) proposed a multi-UAV coordinated vehicular network to analyze driving behavior for improving traffic safety. However, their work is only applicable in scenarios more than two UAVs are available, which is not common at present.

In the application area of risk assessment, UAVs has several advantages: First, UAVs are portable, flexible and robust. Traditional video data collection by cameras mounted on tall physical structures has several limitations including restrictions of the field of view posed by the height of the camera and camera tilt angle. These impair accuracy in tracking the trajectories of the ground vehicles being monitored. Also, the time-consuming and labor-intensive installation process of mounting camera on tall buildings, prohibits timely implementation of ground traffic monitoring. UAVs offer a convenient means to address these limitations as it is possible to easily and quickly dispatch them to the site of interest, and to adjust their spatial locations and camera positions. Secondly, the flexible nature of UAV operations is such that they can facilitate macroscopic and microscopic characterization and analysis of the traffic stream. UAV connectivity to vehicles, infrastructure, and pedestrians can enable intelligent and real-time communications. This capability is useful for safe and efficient CAV operations. Therefore, the use of UAVs for traffic monitoring is promising not only in the current era but also in the prospective era of CAV operations.

Against this background, the objectives of the current study are: (a) propose a framework to use UAVs and V2X connectivity to track the movement of road users and assess potential collisions at intersections, and (2) demonstrate the framework using a case study at an intersection. The developed model, facilitated by advanced machine-learning based models, is intended to enhance the reliable extraction and analysis of trajectory data and detection of collision risk. The proposed model can help traffic engineers assess safety conditions at intersections, recognize root causes of safety hazards, and therefore propose efficient safety countermeasures, and design intersections for improvement not only in the current era but also in the prospective era of CAVs.

## RELATED WORK

In the previous research, several studies have been conducted to investigate UAV applications in road safety management. A few researchers have proposed frameworks to drones photographs to reconstruct accident scenes. Others have compared the use of UAVs to other alternatives for traffic monitoring including manned drone, helicopter, and road patrol vehicles (*12*) and have carried out multiple criteria analysis to identify the most cost-effective monitoring platform. They found that UAV has a lower cost compared to helicopters and is quicker to deploy compared to road patrols, and concluded that UAV is the best option for incident management. Also, there has been research on methods to process the image from drones captured at different shooting angles and altitudes. Ardestani et al. (*13*) prototyped an UAV system for 3D reconstruction of accident scenes. Another researcher (*14*) developed an UAV-based mapping system to acquire scene diagrams, and assessed the reconstruction quality of the images using peak signal-to-noise ratio and structural similarity. As UAV technology continues to develop, research attention is turning towards improving the quality of reconstructed scenes and exploring diverse downstream tasks. For example, Perez et al. (*15*) proposed a low cost and simple method that uses UAV photogrammetry to reconstruct traffic accident scenes. Relatively few research efforts have addressed the UAV applications in safety risk assessment. Risk assessment entails a detailed analysis of vehicle trajectories extracted from UAV-based videos. From the trajectories, potential conflicts, high-risk lanes, and risky maneuvers can be identified and their occurrence predicted. A past researcher (*16*) developed an analysis framework to investigate crash risk at freeway interchange merging areas using data exported from a UAV, and incorporated a driver behavior model to identify factors related to risky driving behavior. Ke et al. conducted a series of research projects to explore UAV applications in smart transportation (*17,18*), and investigated traffic parameter extraction, driver behavior analysis and congestion detection using trajectories from the optical flow model.

  The task of accurately extracting trajectories is probably the most challenging aspect of the UAV-based risk assessment process. In this multi-object tracking (MOT) task, challenges that are encountered include frequent occlusions, initialization and termination of tracks, similarity of appearance, and interactions among different objects. In recent years, the rapid development of convolutional neural-network deep-learning based MOT algorithms with high computing speed and accuracy have been proposed to facilitate the task. Most existing MOT works can be placed into one of two categories: Detection-Based Tracking (DBT) and Detection-Free Tracking (DFT). In DBT models, objects are first detected and then linked into trajectories. In recent tracking studies (*19,20*) benchmarks have been established for DBT models. Bose et al. (*19*) proposed a framework for detecting and tracking multiple interacting objects with due attention paid to dual problems of fragmentation. In their experiments, 89 out of 94 moving objects were correctly tracked and 762 merges and splits were detected. DFT models, on the other hand, are free of pre-trained object detectors but require manual initialization of a fixed number of objects in the first frame (*21,22*). It has been realized by at least one researcher (*23*) that the simultaneous detection and tracking can be carried out using a detection model. DFT models attract significant research attention because they can address disappearing or new objects in the image frame (*24*).

## METHODOLOGY
The proposed framework contains three main stages (Figure 1). The first addresses trajectory extraction and the second stage performs risk assessment. In the first stage, the CenterTrack model (*23*) trained using UAV captured traffic videos is applied in order to obtain real-time and historical trajectories of each road user. In the second and third stage, the crash risk associated with each road user is determined. The scale of the frames and speed of every road user is first calculated using results from the first stage. The crash risk between each pair of tracked road user is then estimated by calculating the time-to-collision (TTC) between them. Implementation details and further discussions are illustrated in subsequent sub-sections of this section of the paper.

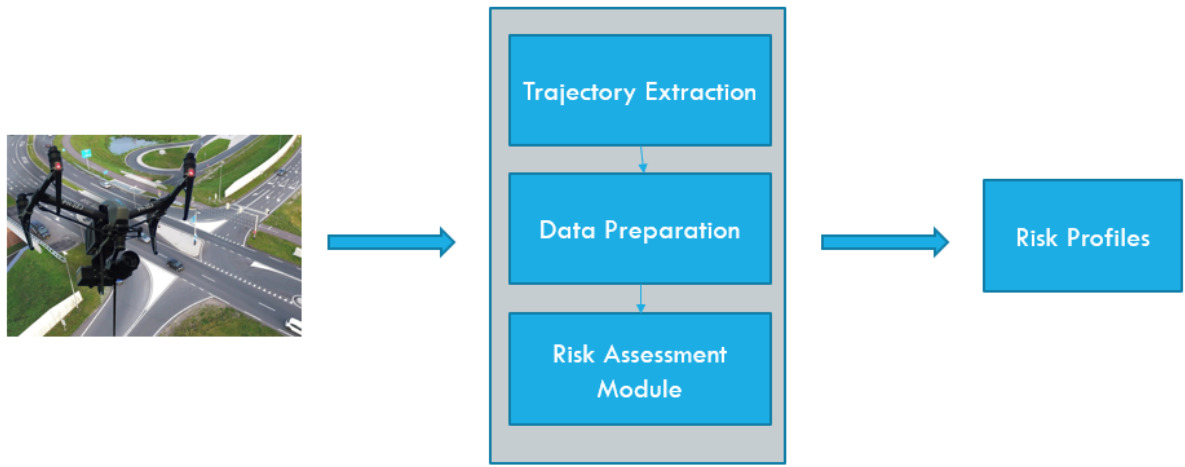

**Figure 1 Overview of the proposed framework**

*Trajectory Tracking*

Accurate trajectory tracking is a key requirement of effective generation of risk profiles. Given an input video sequence, multi-object tracking (MOT) task is required to locate multiple objects, maintain their identities, and yield their individual trajectories. In our scenario, the objects refer to road users at an intersection and the volume of objects is typically large. In addition, given the dynamic traffic pattern, we require a MOT model to capture trajectories of road users quickly and accurately. Recently, convolutional neural network (CNN) based multi object tracking algorithms have been proven as promising and powerful approaches. Zhou et al. in (*23*) developed a CenterTrack model, which is a simultaneous detection and tracking algorithm that is simpler, faster, and more accurate than the state of the art and therefore is a perfect fit to our scenario. Therefore, we developed our method as a further enhancement of the CenterTrack model. CenterTrack identifies each object through its center point and then regresses to a height and width of the object's bounding box. Specifically, it produces a low-resolution heatmap and a size map. The structure of CenterTrack is presented in figure 2. At time t, we are given an image of the current frame and the previous frame, as well as the heatmap of tracked objects in the previous frame. The heatmap presents the distribution of the confidence score of object centers. The heatmap and frames first go through convolutional layers separately, and then are concatenated together to feed into another sequence of convolutional layers. The output from the entire network includes the displacement prediction of objects, height and width of bounding boxes, and a heatmap for the current frame. The tracking task is carried out at a very fast pace since no separate detection is required.

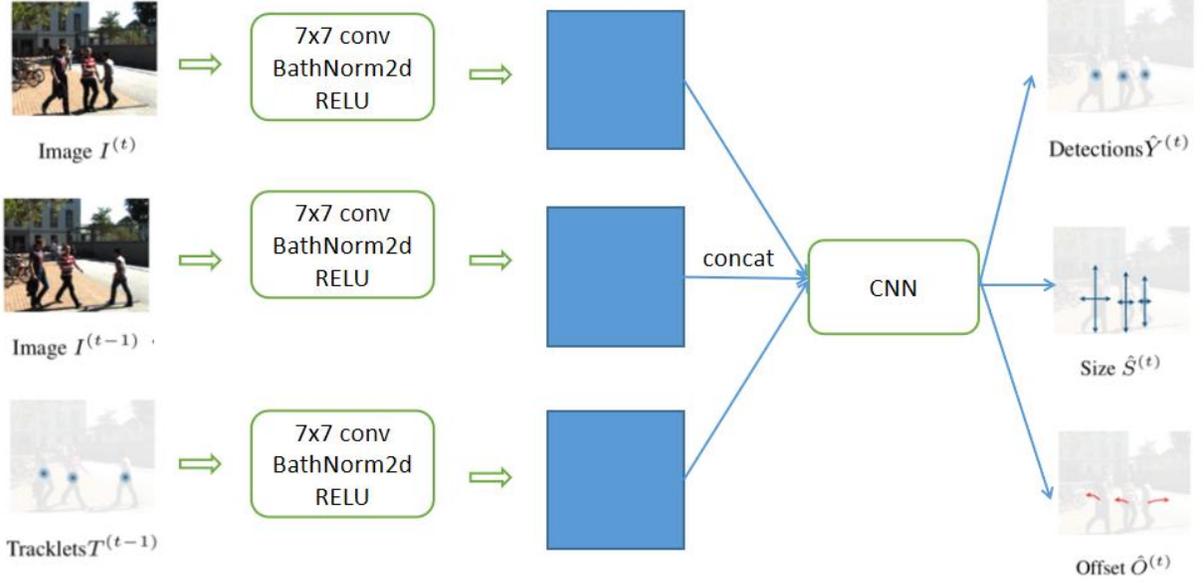

**Figure 2 Structure of the CenterTrack model**

*Data Preparation and Risk Assessment*

The crash risk of road users can be evaluated using the extracted trajectory. First, the data is prepared to obtain the scale of frames and speed of road users. We assume that the length of a general vehicle is 4 meters, and the width is 1.7 meters. A scale can be obtained by aligning detection boxes of vehicles in the video sequence with real dimension of vehicles. The speed of road users is calculated using equation below:

$$v(t) = \frac{\sqrt[2]{(x_t - x_{t-\Delta_t})^2 + (y_t - y_{t-\Delta_t})^2}}{\Delta_t} * scale \ (\frac{m}{s}) \qquad (1)$$

where $\Delta_t$ is the video frame frequency.

After the data preparation, data from each road user is assigned a unique ID. The data includes the center of its bounding box, height and width of its bounding box, speed, and the category it belongs to in every frame. Table 1 presents a summarized set of data (and their notations) used for the risk assessment. We adopted a widely used risk assessment parameter, the time-to-collision (TTC), as a measure of risk. The idea of computing TTC was first brought up by Hayward in 1972 (*25*). The initial definition of TTC is "the time required for two vehicles to collide if they continue at their present speed and on the same path". A lower TTC value corresponds to higher conflict severities and a TTC smaller than 2.5 seconds is typically taken as critical (*26*). Hence, TTC is generally perceived to be a primary and efficient measure in traffic safety assessment. In our study, for any two objects (e.g., object 1 and object 2 in figure 3), the TTC is calculated using Equations (2)-(6) below:

Relative speed:

$$\widehat{v_{rela}} = \widehat{v_2} - \widehat{v_1} \quad (2)$$

$$|v_{rela}| = \sqrt[2]{v_1{}^2 + v_2{}^2 - 2|v_1||v_2|cos\alpha} \quad (3)$$

Distance: $l = \sqrt[2]{(x_2 - x_1)^2 + (y_2 - y_1)^2} \quad (4)$

Projected speed: $v_{projected} = \widehat{v_{rela}} * cos\theta \quad (5)$

TTC: $ttc = \frac{l}{|v_{rela}|} \quad (6)$

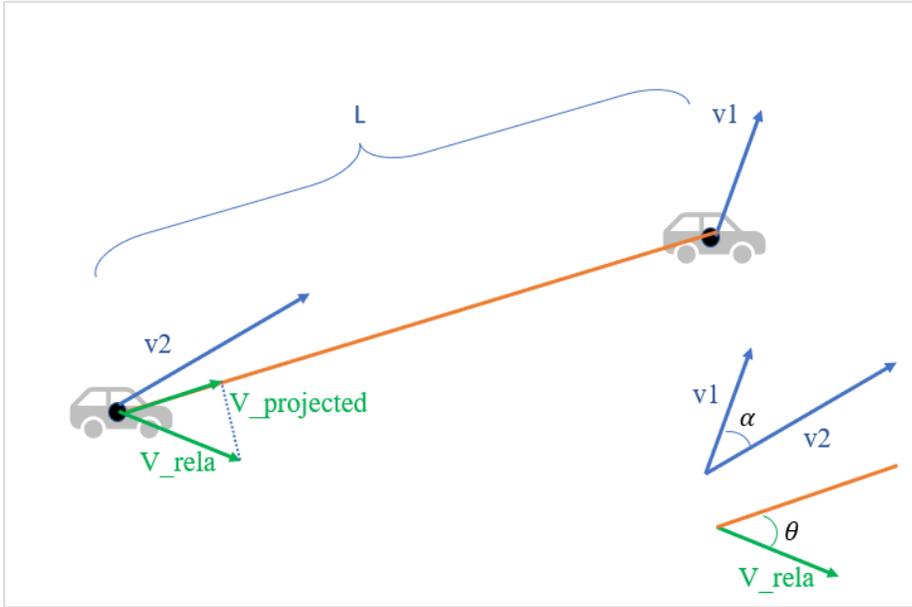

**Figure 3 TTC calculation**

**TABLE 1 Data for risk assessment**

| Data | Notes |
|---|---|
| Across the entire video sequence: | |
| Scale | Match the video to real-word scales |
| Categories | Log all categories of different road users |
| For every individual road users tracked: | |
| $x_t, y_t$ | Location of the center of the bounding box at time t |
| $H_t$ | Height of the center of the bounding box at time t |
| $W_t$ | Width of the center of the bounding box at time t |
| $V_t$ | Speed at time t |
| $C_t$ | Category in which the road user is detected at time t |

From the proposed model, TTC between all tracked road users can be easily achieved and road safety assessment can be done at both in macroscopic and microscopic levels. From the macroscopic perspective, a risk profile of the studied area at every time step can be established by identifying road users suffering from critical TTC. From the microscopic perspective, an individual road user could be informed about TTC regarding neighbors and take actions in an intelligent manner. A detailed demonstration of the assessment is presented in the Case Study section of this paper.

*Performance Evaluation*

The success of the proposed framework is determined by how many risky TTCs it can correctly detect, which further depends on the accuracy of the trajectory tracking task. The ground truth of TTC is obtained by feeding the true trajectory data into the Risk Assessment module. To better fit the UAV scenario, when training the trajectory tracking model, we used video clips provided by VisDrone (*27*), which consists of 56 video clips with 24,198 frames captured by UAVs. The trained model is tested on a test set containing 16 video clips with 6,322 frames. In our work, we consider six categories of road users: pedestrian, bicycle, car, van, truck, bus, motor. We used Multi Object Tracking Accuracy (MOTA) to evaluate the tracking results and detailed calculation methods for MOTA can be found in (*28*). As shown in Table 2, the tracking algorithm gives 64.89 MOTA on the training set and 63.12 MOTA on the testing set.

When using the extracted trajectories to produce risk profile of a studied area, we logged the true positive rate and false negative rate as evaluation matrix. Using 2.5 seconds as a threshold, the TTC between each pair of road users is labeled as: safe vs. critical. A true positive case means both ground truth and our proposed framework detect the TTC between each pair of road users as critical. A false positive case means our proposed framework indicates a TTC as critical while the ground truth shows it is safe. A true negative case refers to situations where both ground truth and our proposed framework indicate that the TTC is safe. Similarly, a false negative case means that our proposed framework gives a safe TTC while the TTC is critical in the ground truth dataset. As shown in Table 3, our model yields a true positive rate of 80% and a false negative rate of 31%. For all the detected TTCs, 78.2% of the model results falls into the ground-truth categories of critical or safe designations.

**TABLE 2 Evaluation of trajectory tracking model**

| MOTA | Dataset |
|------|---------|
| Train set | 64.89 |
| Test Set | 63.12 |

**TABLE 3 Evaluation of risk assessment model**

| Accuracy (=(no. of true positive cases +no. of true negative cases)/no. all cases) | 78.2% |
|------|------|
| True Positive Rate | 80% |
| False Positive Rate | 31% |

## CASE STUDY AND ANALYSIS

To illustrate the analysis framework of the UAV-based risk assessment proposed in sections above, a case study was conducted using drone images captured at an intersection in Tianjin, China. As shown in Figure 4, this intersection has traffic flowing from 4 legs. The video data is provided by an open source dataset (*27*) that includes intersection videos taken under various conditions including sunny weather, good light, and no electromagnetic interference, which could output stable video pictures at a vertical angle. The movements and interactions between vehicles in this intersection were clearly captured at a frame frequency of 30fps.

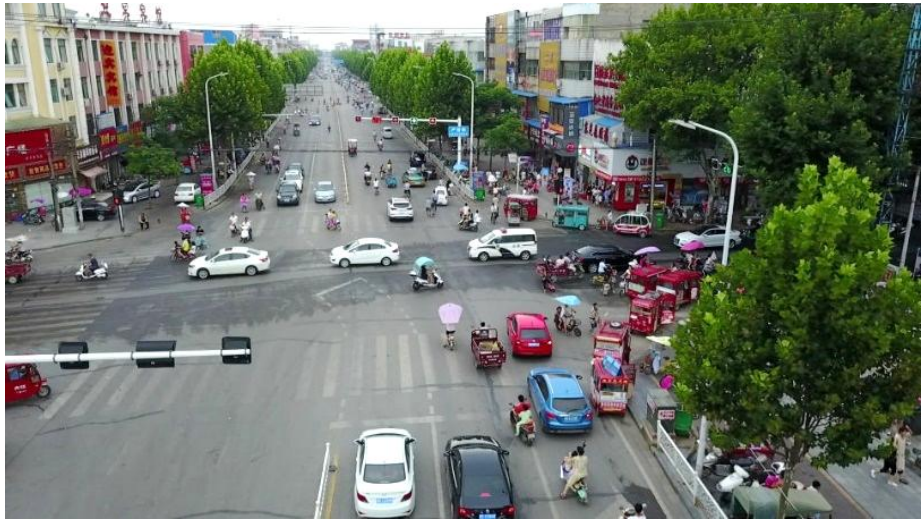

**Figure 4 The intersection used in the case study**

*Risk Profiles*

At each time step, the trajectories of all studied road users were tracked using the deep-learning based model discussed above. Then, the crash risk for every pair of road users was estimated using the TTC equation provided as Eqs (2)-(6). It may be noted that that only positive TTCs are considered in this study and TTCs smaller than 2.5s are labeled as critical. For any road user, if the minimum correlated TTC is critical, the road user is identified as risky. Figure 5 presents a series of consecutive frames where risky road users are highlighted by their bounding box and the number on top of the box is the value of the most critical TTC value correlated the road user in the box. The individual vehicle is indicated by a red circled in the figure. Other road users that are not "safe" are labeled using blue boxes. The number indicated above the boxes is the TTC value between the studied vehicle and the road user in the box. As indicated by Figure 5, dynamic variation of the intersection risk profile is captured by videos. In addition to vehicles, pedestrians and bicycles that are not "safe" are also identified by the proposed framework. Also, the microscopic risk profile can be obtained by extracting information for an individual road user. Figure 6 presents the risk profile of an individual vehicle, and highlights the neighbors that have a "critical" level of TTC with respect to the individual vehicle in question.

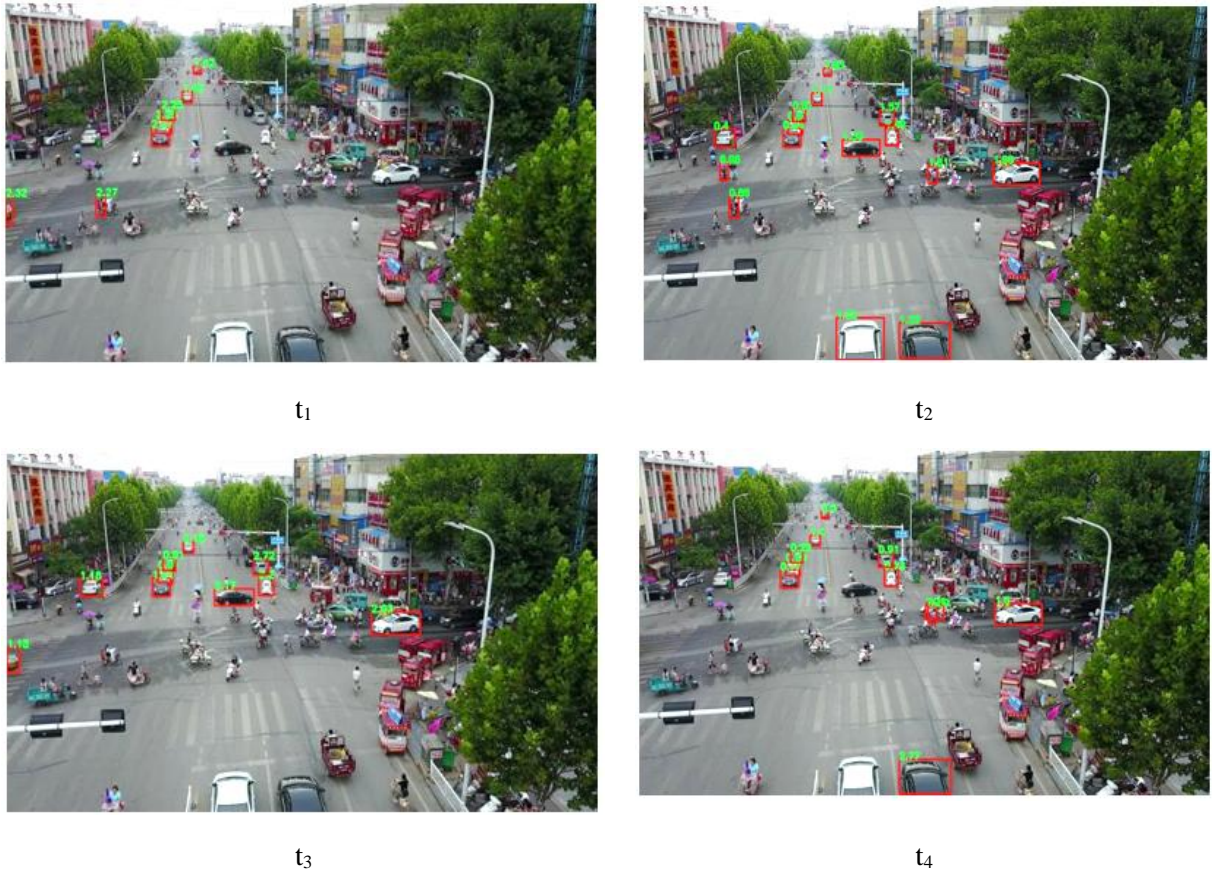

Figure 4 Macroscope risk profile (Dangerous road users are labelled in red boxes. The number on top of boxes are their smallest TTC values)

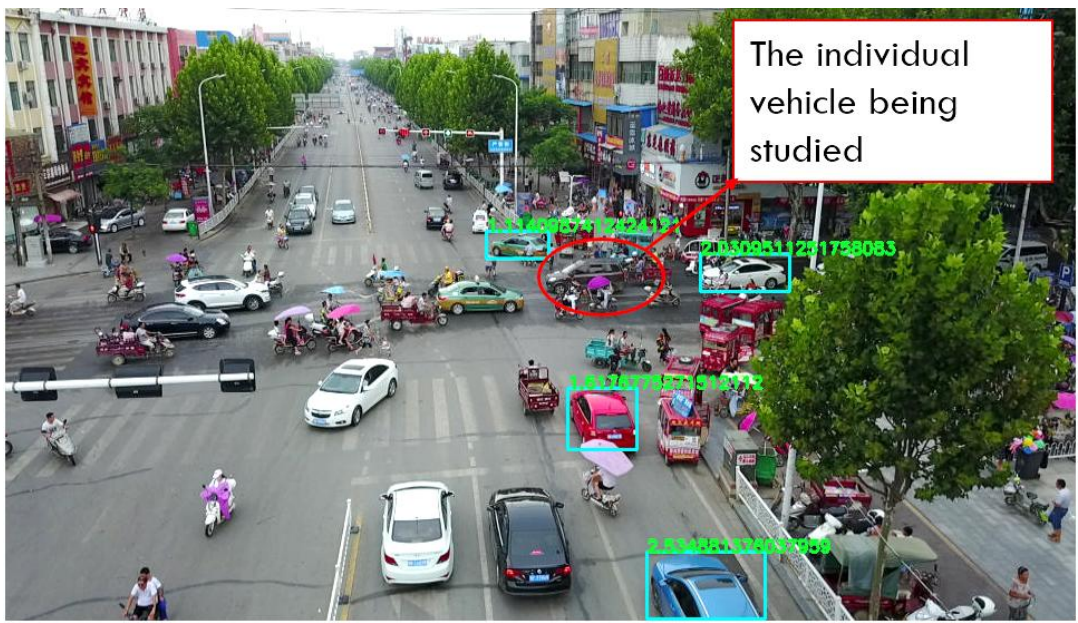

Figure 5 Microscope risk profile of an individual vehicle (circled red)

The benefits of such risk profiles are twofold: First, UAVs can transmit the microscopic risk profile to an individual road user so that the road user would be aware of potential crashes in its surroundings. This is considered particularly conducive to autonomous vehicles because the in-vehicle detectors including cameras and Lidar may fail to identify all potential crashes or hazardous situations due to their narrow detection range (*29-31*). For this, UAVs not only serve as a robust source of information relating to a broad view of the surroundings and a global spatial analysis of the environment, but also accurate data regarding potential risks. Second, UAV can provide a macroscopic risk profile that provides useful insights to urban planners and transportation managers. The risk profile patterns can be identified by summarizing data from the same intersection. In the studied area, 72% of potential collisions are formed between vehicles. In the remaining 28% of potential collisions (Figure 6), 40% are caused by pedestrians and vehicles and 30% collisions happen between trucks and cars. The road agency overseeing the operations of the intersection may be interested in investigating the reasons why risk occur so often between pedestrians and cars, and probably recommend the construction of pedestrian dedicated facilities (special lanes, overhead bridge, tunnel, etc.) to mitigate this problems  In Figure 7, we logged all locations where safety "critical" road user interactions are prevalent. It can be found that most critical potential crashes occurred in the upper right corner of the intersection. This may be due to the large number of bicycles and pedestrians who typically occupy that area where they share the lane with vehicles. As a result, it is difficult for vehicles undertake safe turning maneuvers. Intersection designers could also use the results of such analysis as a basis to carry out intersection improvements.

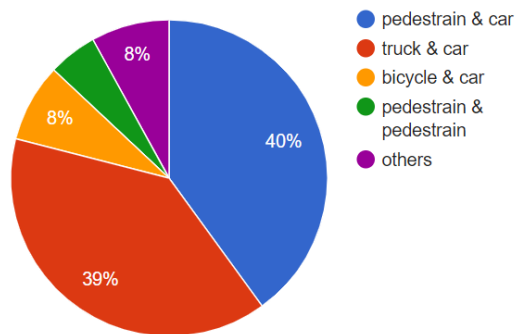

**Figure 6 Road user pairs associated with potential collisions (excludes car-car pairs)**

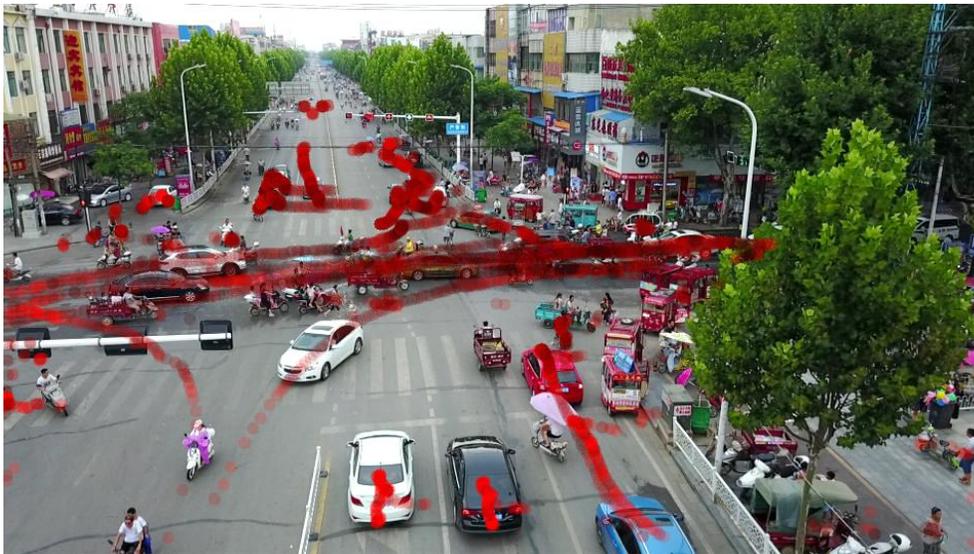

**Figure 7 Summary of all locations where potential collisions occur. A deeper color indicates a less safe interaction (i.e., a smaller TTC) between road user pairs**

*Risk Prediction*

In assessing the risk associated with an intersection, it is also of interest to predict the risk of a vehicle at subsequent time steps. To predict such future crash risk, we deployed a Random Forest classifier, which is a supervised learning algorithm. Random forests create decision trees on randomly selected data samples, obtain predictions from each tree and select the best solution through voting. The major advantage of Random Forest is that it provides an indicator of the feature importance, which offers insights on features that distinguish data samples most. Such information facilitates the identification (before a crash occurs) of vehicles associated with "critical" interactions so that risk could mitigated in a proactive manner. It may be noted that only cars are included as studied objects. This is because the moving pattern of car-related pairs (that is, cars and other road users) is different. The features fed into the random forest model include speed, location, safety condition of the studied vehicle and its neighbors, together with TTC and distance between them in the five previous time steps. The output of the classifier is either "safe" and "risky". Safe means that for the studied vehicle, the smallest predicted TTC at the next time step is greater than 2.5 seconds, and risky means the smallest predicted TTC is less than 2.5 seconds. From the Random Forest model, we obtain the importance of different features regarding future risk prediction, which is calculated by a Gini Importance value that sums over the number of splits (across all tress) that include the feature, proportionally to the number of samples it splits. As shown in figure 8, the top 5 important features and their importance are listed. A "dangerous road user" refers to the neighbor with the smallest TTC with respect to the studied vehicle in the last time step. Features with higher importance contributes more when predicting risky vehicles, indicating we can observe these features to predict the future risk of a vehicle. The threshold of these features could also be extracted from Random Forest classifier. The threshold values are not discussed in this paper because the threshold values are very specific to the studied area and the time of capturing the video, and cannot be generalized. According to the Random Forest classifier, the vehicle speeds in the previous time steps are most related to its future safety condition. The status of the vehicle's neighbors, particularly the location and speed of its dangerous neighbor, also plays a significant role in deterring a vehicle's future safety condition. In practice, warning messages could be generated based on these results and sent to the vehicles concerned, to remind them to be aware of imminent danger of traffic collision. In addition, the location of a vehicle plays a key role in identifying its risk, which aligns with our finding in the section above that a critical region exist in an intersection. The critical region may result from improper design of the traffic signal phases, lanes or configurations of the intersection. The results of this analysis can serve as a basis for giving special attention should be given to such intersection to improve their safety.

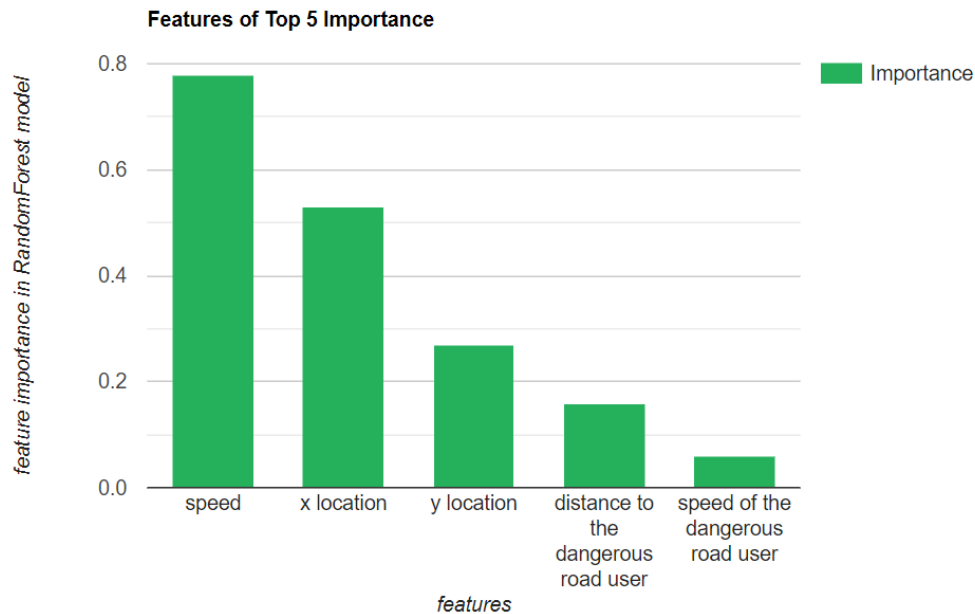

**Figure 8 Features of top 5 importance in the Random Forest model**

## CONCLUSION

This study presents a methodology to assess the traffic safety of an intersection utilizing UAV video. The methodology explores the potential crashes between each pair of road users by extracting their trajectories from video images and calculating their time-to-collision values. To develop the trajectories of all detected road users at an efficient and accurate manner, the study used a deep-learning based multi-object tracking algorithm. The trajectory data were packed and re-scaled for the risk assessment. Then, a method was suggested to compute the crash risk between pair of road users by calculating the time-to-collision between them. Based on the TTC value, "dangerous" road users whose smallest TTC is less than a threshold (2.5s in our study) are identified and a macroscopic risk profile can be established and presented to road agency that manage the intersection. An individual road user can acquire its own microscopic risk profile from the UAV so that it can make its safe and informed movement decisions accordingly. In the case study, we demonstrated how our framework could assist intersection management in the current era and in the future era of autonomous driving. The results showed that by investigating consecutive macroscopic risk profiles, spatial-temporal pattern of risk profiles can be observed. A critical zone where potential collisions happen most frequently and the most dangerous road user categories could be identified. Urban planners and intersection managers may find these results useful in their efforts to improve the traffic control or configuration at intersections. In addition, we deployed a Random Forest model to predict the safety condition of a vehicle by utilizing historical risk profiles. It was concluded that the travel speed is the most critical parameter that determines a vehicle's future safety condition. The speed of the vehicle's neighbor also contributes to the vehicle's future safety status. With the proposed model, traffic engineers can be placed in a better position to propose efficient countermeasures to enhance road safety. Also, the proposed model can provide CAVs information that is helpful for making informed driving decisions and make available data for traffic engineers that may be considering making intersection improvements.

**ACKNOWLEDGMENTS**

This work was supported by Purdue University's Center for Connected and Automated Transportation (CCAT), a part of the larger CCAT consortium, a USDOT Region 5 University Transportation Center funded by the U.S. Department of Transportation, Award #69A3551747105. The contents of this paper reflect the views of the authors, who are responsible for the facts and the accuracy of the data presented herein, and do not necessarily reflect the official views or policies of the sponsoring organization. This manuscript is herein submitted for PRESENTATION ONLY at the 2022 Annual Meeting of the Transportation Research Board.

**AUTHOR CONTRIBUTIONS**

The authors confirm contribution to the paper as follows: all authors contributed to all sections. All authors reviewed the results and approved the final version of the manuscript.

**REFERENCES**

1. Prevot, T., Rios, J., Kopardekar, P., Robinson III, J. E., Johnson, M.,& Jung, J. (2016). UAS traffic management (UTM) concept of operations to safely enable low altitude flight operations. In 16th AIAA Aviation Technology, Integration, and Operations Conference, (June), 1–16. https://doi.org/10.2514/6.2016-3292.

2. MENA (2016). ''United States: Global market for commercial applications of drone technology valued at over $127 Bn.'' MENA Report, May, 2016.

3. Nath, T. (2015). ''How drones are changing the business world'', Investopedia. Retrieved December 25, 2018, from https://www.investopedia.com/articles/investing/010615/how-drones-are-changing business-world.asp.

4. Outay, Fatma, Mengash, Hanan Abdullah, and Adnan, Muhammad. "Applications of Unmanned Aerial Vehicle (UAV) in Road Safety, Traffic and Highway Infrastructure Management: Recent Advances and Challenges." Transportation Research. Part A, Policy and Practice 141 (2020): 116-29. Web.

5. Kure, M., 2020. How drones become a valuable tool for the auto insurance industry. Forbes article, Accessed on May 2020, https://www.forbes.com/sites/sap/2020/01/29/how-drones-become-a-valuable-tool-for-the-auto-insurance-industry/#56ff18a61ac9.

6. Zhang, H., Liptrott, M., Bessis, N., Cheng, J., 2019, September. Real-time traffic analysis using deep learning techniques and UAV based video. In: 2019 16th IEEE International Conference on Advanced Video and Signal Based Surveillance (AVSS). IEEE, pp. 1–5.

7. Ke, R., Li, Z., Kim, S., Ash, J., Cui, Z., Wang, Y., 2017. Real-time bidirectional traffic flow parameter estimation from aerial videos. IEEE Trans. Intell. Transp. Syst. 18 (4), 890–901.

8. Raj, C.V., Sree, B.N., Madhavan, R., 2017, July. Vision based accident vehicle identification and scene investigation. In: 2017 IEEE Region 10 Symposium (TENSYMP). IEEE, pp. 1–5.

9. Zink, J., Lovelace, B., 2015. Unmanned aerial vehicle bridge inspection demonstration project. Research Project. Final Report, 40. Accessed on May 2020 http://www. dot.state.mn.us/research/TS/2015/201540.pdf.

10. Kim, N.V., Chervonenkis, M.A., 2015. Situation control of unmanned aerial vehicles for road traffic monitoring. Modern Appl. Sci. 9 (5), 1.

11. Sharma, V., Chen, H-C., Kumar, R., 2017. Driver behavior detection and vehicle rating using multi-UAV coordinated vehicular networks, J. Comput. Syst. Sci., 86, 3–32.

12. Mehmood, S., Ahmed, S., Kristensen, A. S., Ahsan, D., 2018, May. Multi Criteria Decision Analysis (MCDA) of Unmanned Aerial Vehicles (UAVs) as a Part of Standard Response to Emergencies. In: 4th International Conference on Green Computing and Engineering Technologies; Niels Bohrs Vej 8, Esbjerg, Denmark.

13. Ardestani, S.M., Jin, P.J., Volkmann, O., Gong, J., Zhou, Z., Feeley, C., 2016. 3D Accident Site Reconstruction Using Unmanned Aerial Vehicles (UAV). In: Presented in 95th Annual Meeting, Transportation Research Board, Washington DC, USA. Paper No. 16-5703.

14. Su, S., Liu, W., Li, K., Yang, G., Feng, C., Ming, J., Yin, Z., 2016. Developing an unmanned aerial vehicle-based rapid mapping system for traffic accident investigation.
Aust. J. Forensic Sci. 48 (4), 454–468.

15. Perez, J.A., Gonçalves, G.R., Rangel, J.M.G., Ortega, P.F., 2019. Accuracy and effectiveness of orthophotos obtained from low cost UASs video imagery for traffic accident scenes documentation. Adv. Eng. Software 132, 47–54. Cooner, S.A., Balke, K.N., 2000.

16. Gu, Xin, Abdel-Aty, Mohamed, Xiang, Qiaojun, Cai, Qing, and Yuan, Jinghui. "Utilizing UAV Video Data for In-depth Analysis of Drivers' Crash Risk at Interchange Merging Areas." Accident Analysis and Prevention 123 (2019): 159-69. Web.

17. Ke, R., Li, Z., Kim, S., Ash, J., Cui, Z., Wang, Y., 2017. Real-time bidirectional traffic flow parameter estimation from aerial videos. IEEE Trans. Intell. Transp. Syst. 18 (4), 890–901.

18. Ke, R., Li, Z., Tang, J., Pan, Z., Wang, Y., 2018. Real-time traffic flow parameter estimation from UAV video based on ensemble classifier and optical flow. IEEE Trans. Intell. Transp. Syst. 20 (1), 54–64.
Ke, R., Feng, S., Cui, Z., Wang, Y., 202

19. B. Bose, X. Wang, and E. Grimson, "Multi-class object tracking algorithm that handles fragmentation and grouping," in Proc.IEEE Comput. Soc. Conf. Comput. Vis. Pattern Recognit., 2007, pp.1–8.

20. B. Song, T.-Y. Jeng, E. Staudt, and A. K. Roy-Chowdhury, "A stochastic graph evolution framework for robust multi-target tracking," in Proc. Eur. Conf. Comput. Vis., 2010, pp. 605–619.

21. W. Hu, X. Li, W. Luo, X. Zhang, S. Maybank, and Z. Zhang, "Single and multiple object tracking using log-euclidean riemannian subspace and block-division appearance model," IEEE Trans. Pattern Anal. Mach. Intel., vol. 34, no. 12, pp. 2420–2440, Dec. 2012.

22. L. Zhang and L. van der Maaten, "Structure preserving object tracking," in Proc. IEEE Comput. Soc. Conf. Comput. Vis. Pattern Recognit., 2013, pp. 1838–1845.

23. Zhou, Xingyi, Koltun, Vladlen, and Krähenbühl, Philipp. "Tracking Objects as Points." Computer Vision – ECCV 2020. Cham: Springer International, 2020. 474-90. Lecture Notes in Computer Science. Web.

24. Luo, Wenhan, Xing, Junliang, Milan, Anton, Zhang, Xiaoqin, Liu, Wei, Zhao, Xiaowei, and Kim, Tae-Kyun. (2014). Web.

25. Hayward, J.C., 1972. Near-miss determination through use of a scale of danger. Highway Research Record, 384, 24-34.

26. Nadimi, Navid, NaserAlavi, Seyed Saber, and Asadamraji, Morteza. "Calculating Dynamic Thresholds for Critical Time to Collision as a Safety Measure." Proceedings of the Institution of Civil Engineers. Transport (2020): 1-10. Web.

27. Pengfei Zhu, Longyin Wen, Dawei Du, Xiao Bian, Qinghua Hu, Haibin Ling. Vision Meets Drones: Past, Present and Future. arXiv preprint arXiv:2001.06303 (2020).

28. Bernardin, Keni, and Stiefelhagen, Rainer. "Evaluating Multiple Object Tracking Performance: The CLEAR MOT Metrics." EURASIP Journal on Image and Video Processing 2008.1 (2008): 1-10. Web.

29. Chen, Sikai, Jiqian Dong, Paul Ha, Yujie Li, and Samuel Labi. "Graph neural network and reinforcement learning for multi-agent cooperative control of connected autonomous vehicles." Computer‐Aided Civil and Infrastructure Engineering 36, no. 7 (2021): 838-857.

30. Dong, Jiqian, Sikai Chen, Yujie Li, Runjia Du, Aaron Steinfeld, and Samuel Labi. "Space-weighted information fusion using deep reinforcement learning: The context of tactical control of lane-changing autonomous vehicles and connectivity range assessment." Transportation Research Part C: Emerging Technologies 128 (2021): 103192.

31. Ha, Paul Young Joun, Sikai Chen, Runjia Du, Jiqian Dong, Yujie Li, and Samuel Labi. "Vehicle connectivity and automation: a sibling relationship." Frontiers in Built Environment 6 (2020): 199.